\documentclass{article}
\usepackage[table]{xcolor}
\usepackage[utf8]{inputenc}
\usepackage[margin=2cm]{geometry}
\usepackage[table]{xcolor}

\usepackage{algorithm}
\usepackage[noend]{algpseudocode}
\usepackage{comment}

\newcommand{\emilie}[1]{\textcolor{magenta}{[{\em EMILIE SAYS}: #1]}}

\title{DeepVir - Graphical Deep Matrix Factorization for \emph{In Silico} Antiviral Repositioning: Application to COVID-19}

\author{Aanchal Mongia$^{1}$, Stuti Jain$^{3}$, Emilie Chouzenoux$^{2*}$ \& Angshul Majumdar$^{3*}$\\ \\
        $^{1}$Dept. of CSE, IIIT - Delhi, India, 110020\\
	  	$^{2}$CVN, Inria Saclay, Univ. Paris Saclay, 91190 Gif-sur-Yvette, France\\
       	$^{3}$Dept. of ECE, IIIT - Delhi, India, 110020\\
        $^{*}$Corresponding authors contact/Email:\\emilie.chouzenoux@centralesupelec.fr,  angshul@iiitd.ac.in\\}

\usepackage{graphicx}

\usepackage{url}
\usepackage{comment}
\usepackage{mathtools}
\usepackage{tikz}

\usepackage{makecell}
\usepackage{soul}

\begin{document}

\maketitle



\begin{abstract}
This work formulates antiviral repositioning as a matrix completion problem where the antiviral drugs are along the rows and the viruses along the columns. The input matrix is partially filled, with ones in positions where the antiviral has been known to be effective against a virus. The curated metadata for antivirals (chemical structure and pathways) and viruses (genomic structure and symptoms) is encoded into our matrix completion framework as graph Laplacian regularization. We then frame the resulting multiple graph regularized matrix completion problem as deep matrix factorization. This is solved by using a novel optimization method called HyPALM (Hybrid Proximal Alternating Linearized Minimization). Results on our curated RNA drug virus association (DVA) dataset shows that the proposed approach excels over state-of-the-art graph regularized matrix completion techniques. When applied to \emph{in silico} prediction of antivirals for COVID-19, our approach returns antivirals that are either used for treating patients or are under for trials for the same.  


\textbf{Contact:} angshul@iiitd.ac.in

\end{abstract}

\section{Introduction}

The problem of matrix completion has been deployed in numerous applications of computer science and bioinformatics. One can mention, for instance, recommender systems \cite{cheng2014lorslim}, wireless sensor networks \cite{cheng2012stcdg} and MIMO channel estimation \cite{vlachos2018massive}. Matrix completion also forms the foundation of many biological interaction problems including gene expression imputation \cite{kapur2016gene}, drug-target interaction prediction \cite{survey1,survey2,zheng2020predicting}, miRNA disease association \cite{chen2018predicting,li2017mcmda} and gene disease association \cite{natarajan2014inductive}. As seen in the aforementioned studies, matrix completion is turning out to one of the most promising approaches for modelling biological interactions. 

Matrix completion for \emph{in silico} drug re-positioning is an established topic \cite{survey1,zhang2018predicting}. In such context, the matrix rows relates to drugs while the columns correspond to the diseases. The information about a drug being effective for a disease is encoded with a one into the matrix and zero, otherwise. The matrix is only partially filled since information about all drug disease associations are not known. Matrix completion amounts to filling the missing entries by relying and interpolating over the available knowledge.  

Our recent work~\cite{mongia2020computational} introduced a dataset of antiviral drug virus association (DVA). It shows how the matrix completion framework can be used to computationally predict the drugs that could be effective against SARS-CoV-2 (severe acute respiratory syndrome coronavirus 2), the virus responsible for the ongoing pandemic, COVID-19 (COrona VIrus Disease-2019). In the aforesaid dataset, we manually curated drug-virus associations from \texttt{drugbank.ca}. We collected all the anti-viral drugs proved to be effective against viruses infecting humans along with the similarity information between the viruses (genomic structure-based similarity) and drugs (chemical structure-based similarity). The dataset contains both DNA and RNA viruses known to infect human beings. State-of-the-art off-the-shelf matrix completion techniques were then employed to solve the association prediction problem. It was found that graph regularised techniques, including similarities based on genomic structure of viruses and chemical structure of antiviral drugs, performed better than the non-regularised competitors. The most successful techniques were graph regularized versions of matrix completion \cite{mongia2020drug}, binary matrix completion \cite{mongia2020computationalb} and (shallow) matrix factorization \cite{ezzat2017drug}.

In this present work, we propose to solve the problem of drug-virus association prediction via graph regularized deep matrix factorization. We propose a novel theoretically sounded algorithm for this task. Moreover, since our final goal is to repurpose drugs for COVID-19, an RNA virus, we have pruned our DVA dataset to consists only RNA viruses. DNA viruses are not meaningful in this context \cite{o2016viral}, hence we will not be considering them in this work. Furthermore, in addition to the graph based on chemical structure similarity of drugs \cite{mongia2020computational}, one more similarity graph has been built on the mode of action of the drugs. Similarly, an additional similarity graph on viruses is built, based on the human symptoms caused by the virus' infection. The current work uses the symptom based similarities along with the genomic similarities used in the prior work \cite{mongia2020computational}.  

A prior study on graph regularised deep matrix factorization \cite{mongia2019drug} was based on the alternating direction method of multipliers (ADMM) approach \cite{wang2019global}. A major issue with ADMM is that the convergence guarantees are rather mild in such challenging non convex scenario. Furthermore, ADMM method requires the resolution of costly inner steps as well as the introduction of several intermediary variables that may be detrimental to the practical efficiency of the method. Owing to the aforesaid reasons, we propose a novel resolution scheme called HyPALM (Hybrid Proximal Alternating Linearized Minimization), relying on the recent techniques developed in \cite{RepettiJOGO,bolte2014proximal,Abboud2014}. To summarize the main contributions of our proposed work are as follows:

\begin{itemize}
\item  An original graph structure introducing (i)  symptoms based viral similarity in the virus-virus graph, (ii)  mechanism-of-action based drug similarity in the drug-drug graph, leading to the so-called multi graph regularized deep matrix factorization (GRDMF) model; We call the approch GRDMF throughout this paper.
\item A novel optimization method, named HyPALM, for addressing the resulting matrix factorization  problem with sounded convergence guarantees on its iterates.
\end{itemize}

\section{Results}
\subsection{Algorithm Selection}
In this subsection, we study the choice of technique to solve the GRDMF problem \eqref{eq:mgrdmf1}. In the past work \cite{mongia2020deep,mongia2019drug}, the proposed formulation has been solved using ADMM (alternating direction method of multipliers) \cite{boyd2011alternating,Komodakis}. We introduce here a novel resolution strategy named HyPALM. In contrast with ADMM, the iterates of HyPALM are guaranteed to converge to a stationnary point of the non-convex problem. The HyPALM algorithm steps are depicted in the Methods section of the paper.

We randomly drop 10 \% of association values in the drug-virus association matrix and report the AUC (Area under the ROC curve) and AUPR (Area under the precision recall curve) on the test entries (averaged over 10 runs of cross validation) after solving 2-layer and 3-layer GRDMF problem with ADMM (first two columns) and HyPALM (last two columns). As can be seen, HyPALM solution gives much better prediction results,  for both 2 and 3 layers, and both evaluation metrics . Hence, owing to the better performance and the assessed convergence guarantees, we propose to retain HyPALM in the remaining of the paper for our simulations with the GRDMF model.

\begin{table}[h!]
\centering
\begin{tabular}{l|l|l||l|l}
\hline \hline
Model &   GRDMF-2L    & GRDMF-3L &  GRDMF-2L     & GRDMF-3L  \\
Algorithm  & ADMM  &  ADMM  &  HyPALM   &  HyPALM\\
\hline
AUC       & 0.9076 	& 0.9143 & {0.9457}	& {0.9516}		\\
AUPR      & 0.6523	& 0.6945 &	{0.8180}	& {0.8038}	 \\
\hline 
\end{tabular}

\caption{\textbf{AUC and AUPR obtained after running ADMM and HyPALM algorithms for DLMF matrix completion for drug-virus association prediction.}}
\label{t:ablation}
\end{table}

\subsection{Evaluation}
We compare the performance of our method with other recent graph regularized matrix completion techniques such has graph regularized matrix completion (GRMC) \cite{mongia2020drug}, graph regularized binary matrix completion (GRBMC) \cite{mongia2020computational} and graph regularized techniques like graph regularized matrix factorization (GRMF) \cite{ezzat2017drug}. The authors of \cite{mongia2020computational} show that GRMF and GRBMC perform well in predicting drug-virus association and hence estimating drugs for SARS-CoV-2. We evaluate the performance of 2-layers and 3-layers graph regularized deep matrix factorization model (GRDMF)  with the similarities provided by the method from \cite{mongia2020computational} and by the newly proposed models GRDMF-2L and GRDMF-3L.

To quantitatively measure and compare the performance of the proposed and other benchmark algorithms, we carry out 10-fold cross validation on the drug-virus association dataset in 3 settings and observe the standard metrics such as mean AUC and AUPR. The first setting (CV1) corresponds to randomly hiding 10 \% association entries as the test set in the association matrix, while the other two settings CV2 and CV3 correspond to randomly hiding 10 \% viruses and drugs in the data as test set respectively. Such a validation technique is a standard to evaluate the accuracy in association/interaction prediction problems \cite{survey2,wang2018drug}. As can be observed from Table \ref{t:10fcv}, both version of our algorithm (GRDMF-2L and GRDMF-3L) perform much better than the state-of-the-art in terms of AUPR and gives comparable results in terms of AUC. This metric is more important than AUC because it penalizes more highly ranked false positives so that only highly ranked drug-virus pairs in prediction would be recommended for biological or chemical tests. 
It can be seen that the prediction results do not improve much in the 3-layers version. Hence, we use the 2-layers GRDMF for further experiments, as it presents lower complexity.

\begin{table}[h!]
\centering
\begin{tabular}{l|l|lll|ll}
\hline \hline
&Model   &   GRMC    &GRBMC    &GRMF  & GRDMF-2L &GRDMF-3L\\
\hline
\hline
CV1&AUC       &0.9092	&0.9168	&0.8926& 0.9457	&	 \textbf{0.9516}\\
&AUPR        &0.5022	&0.55	&0.6393& \textbf{0.8180}	& 0.8038\\

\hline 
\hline
CV2&AUC       & 0.7907 &0.7624	&0.7805& \textbf{0.7791}	& 0.7595\\
&AUPR       &0.3304	&0.3076	&0.3461& \textbf{0.4610}	& 0.4541\\
\hline 
\hline
CV3&AUC    &0.8685	&0.8775	&0.8841& 0.8953	&  \textbf{0.9028}\\
&AUPR      & 0.7282	&0.7042	&0.6885& 0.7513	& \textbf{0.7549}\\
\hline
\hline
\end{tabular}
\caption{\textbf{AUC and AUPR for association prediction for all techniques under the 3 cross validation settings with multiple similarities used for drugs and viruses}.}
\label{t:10fcv}
\end{table}

Let us analyse deeper the results in Table 2. The dataset contains more drugs (86) than viruses (23). In CV1 settings, when entries are randomly omitted, without skipping drugs and viruses our method performs the best. There is enough data and hence our method does not overfit. The results improve from one (GRMF) to two and three layers. In the CV2 setting, when viruses are omitted, the size of the dataset reduces. The proposed deeper method overfits and hence performs poorly compared to shallower techniques. CV3 also reduces the size of the dataset by omitting drugs, but since the number of drugs are significantly larger than the number of viruses, the effect is not pronounced on our proposed algorithm and we see an improved prediction. 

We then measure the precision and recall at top $k$ (Pre@k and Rec@k, $k=3$, $k=5$ and $7$) drug recommendations for viruses while implementing Leave-one-out-cross validation (LOOCV) by hiding (i.e., leaving out) the association profile of every virus. Performing LOOCV tells us how well is the algorithm doing in terms for predicting drugs for current known viruses. This is the most realistic setting, where all the drugs and all the viruses (barring the test virus) are considered. One can see that our algorithm gives comparable precision values and exhibits the best recall at all values of $k$. This appears as the most realistic case since, on average, there are 4 to 5 drugs for each virus.

\begin{table}[h!]
\centering
\begin{tabular}{l|l|lll|ll}
\hline \hline
&Model      & GRMC    &GRBMC    &GRMF       & GRDMF-2L &GRDMF-3L\\
\hline
\hline
k=3&Pre@k     &0.3478 & 0.3623 &  0.3623 & 0.3043 & {0.3478}\\
&Rec@k        &0.4204 &0.4421 &0.4639 & 0.4938  & {0.5398} \\
\hline
\hline

k=5&Pre@k     &0.2522	& 0.2609	& 0.2609&  0.2522	& 0.2870\\
&Rec@k         &0.4554	&0.5195	&0.5152& 0.5956	&	 0.6542\\

\hline 
\hline

k=7&Pre@k     &0.2174 &0.2112 & 0.2174 &  0.2050  &  0.2360 \\
&Rec@k        &0.5382 &0.5288 &0.5375 & 0.6272  & 0.6892  \\
\hline
\hline

\hline
\hline
\end{tabular}
\caption{\textbf{Precision@k and Recall@k for association prediction for k=5 and 10 with multiple similarities used for drugs and viruses.}}
\label{t:prerec}
\end{table}

\begin{figure}[H]
    \centering
    \includegraphics[width=0.6\textwidth]{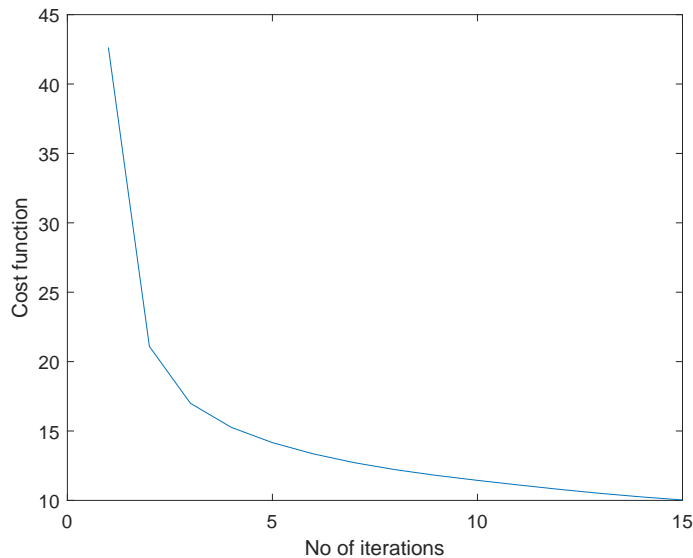}
    \caption{Evolution of the loss function $F$ along the iterations of HyPALM, on the drug-virus association prediction for cross validation setting 1.}
    \label{fig:convergence_plot}
\end{figure}

The values for the GRDMF model parameters $\mu, k_1, k_2, k_3$ and for the HyPALM parameters $\alpha,\vartheta$ have been set by performing cross-validation on the training set (see Supplementary Material for the list of retained values). Moreover, we set the number of iterations $K = 10$ in HyPALM, which appears sufficient to reach stability of the iterates. The time taken by the proposed approach is comparable to the state-of-the art techniques and has been reported in Table \ref{t:time}. All experiments have been performed on a system with Intel(R) Core(TM) i7 x64-based processor and 8 GB RAM. We also illustrate in Figure \ref{fig:convergence_plot} the practical convergence profile of our algorithm HyPALM, on an example. 

\begin{table}[h!]
\centering
\begin{tabular}{l|lll|ll}
\hline \hline
Model   &   GRMC    &GRBMC    &GRMF & GRDMF-2L &GRDMF-3L\\
\hline
\hline
Time   &   4.327 &	0.002 &	0.026  & {0.07} &	{0.07}\\
\hline
\end{tabular}
\caption{\textbf{Execution times (in seconds) for association prediction for all techniques}.}
\label{t:time}
\end{table}

\subsection{Ablation study}
In order to quantitatively assess the necessity/importance of each of the similarity models at use, we conducted an ablation study where we used various combinations of similarity (graph regularizations) matrices within the model to predict 10 \% of the test association matrix, which were randomly hided as test set (cross validation setting CV1). The results have been reported in Table \ref{t:ablation}. 

Here, $S1_{d}$ corresponds to chemical structure based similarity for drugs and $S2_{d}$ corresponds to the similarity based on mechanism of action of drugs. Similarly, $S1_{v}$ corresponds to the genomic structure based similarity and $S2_{v}$ corresponds to symptomatic similarity between the viruses.

\begin{table}[h!]
\centering
\begin{tabular}{l|lllll}
\hline \hline
Similarity   &   ($S1_{d},S1_{v}$)   & ($S1_{d},S2_{v}$)  & ($S2_{d},S2_{v}$)       & ($S2_{d},S1_{v}$) & ($S1_{d}+S2_{d},S1_{v}+S2_{v}$)\\
\hline
AUC       & {0.8897}  & {0.9010}  & {0.9229}  & {0.9289} & {0.9457}\\

AUPR   & {0.6581} & {0.6790} & {0.6970} & {0.6933} & {0.8180)} \\
\hline 
\end{tabular}
\caption{\textbf{AUC and AUPR for association prediction for various combinations of similarities with DMF for all techniques under the cross validation setting CV1}.}
\label{t:ablation}
\end{table}

Let us make a few observations from this study. Firstly, taking into account the new similarities gives the best prediction results. Secondly, when we use the chemical structure based drug similarity ($S1_d$) and vary the viral similarities from $S1_v$ to $S2_v$ (columns 1 and 2), we observe that symptomatic similarity ($S2_v$) yields better results than genomic similarity ($S1_v$) and hence contributes more towards the prediction task. Similar behavior appears when using the genomic similarity ($S1_v$) for viruses and varying the kind of drug similarity used (columns 1 and 4 of Table~\ref{t:ablation}). In other words, our results assess that the new similarities contribute more towards the drug recommendation and greatly improve the prediction results when taken into account in combination with the previous ones. Lastly, we notice that simply replacing the older similarities with the new ones (column 3 of Table~\ref{t:ablation}) improves the results. Hence, this ablation study establishes the benefits of the new similarity models that we introduced.

\subsection{SARS-Cov-2 prediction}

In this subsection, we predict top-10 drugs using the best performing solution  GRDMF-2L and the baseline algorithms for the novel Coronavirus (SARS-CoV-2). We display the results in Table \ref{t:top10-covid}.

\begin{table}[h!]
\centering
\begin{tabular}{l|l|l|l|l}
\hline \hline
Rank&{GRDMF-2L} &GRMC&GRBMC&GRMF\\
\hline 
\hline		
1&\textcolor{green}{Ribavirin}	&\textcolor{green}{Remdesivir}	&\textcolor{green}{Remdesivir}&	\textcolor{green}{Remdesivir}\\

2&\textcolor{green}{Chloroquine}	&\textcolor{green}{Ribavirin}	&\textcolor{green}{Ribavirin}&	\textcolor{green}{Ribavirin}\\

3&\textcolor{green}{Remdesivir}	&\textcolor{green}{Umifenovir}	&\textcolor{green}{Umifenovir}&	\textcolor{green}{Umifenovir}\\

4&\textcolor{green}{Umifenovir}	&\textcolor{blue}{Telaprevir}	&\textcolor{blue}{Telaprevir}&	\textcolor{blue}{Boceprevir}\\

5&\textcolor{green}{Favipiravir}	&\textcolor{blue}{Boceprevir}	&\textcolor{green}{Sofosbuvir}&	\textcolor{blue}{Telaprevir}\\

6&\textcolor{blue}{Baloxavir marboxil}	&\textcolor{red}{Palivizumab}	&\textcolor{blue}{Boceprevir}&	\textcolor{green}{Sofosbuvir}\\

7&\textcolor{green}{Interferon alfa-2a, Recombinant}	&\textcolor{green}{Sofosbuvir}	&\textcolor{red}{Palivizumab}& \textcolor{red}{Palivizumab}\\

8&\textcolor{red}{Geldanamycin}	&\textcolor{blue}{Simeprevir}	&\textcolor{green}{Chloroquine}& \textcolor{blue}{Simeprevir}\\

9&\textcolor{red}{Laninamivir}	&\textcolor{red}{Taribavirin}	&\textcolor{blue}{Simeprevir}&	\textcolor{red}{Taribavirin}\\

10&\textcolor{blue}{Dolutegravir}	&\textcolor{green}{Chloroquine}	&\textcolor{red}{Taribavirin}& \textcolor{blue}{Paritaprevir}\\			
\hline 
\hline
\end{tabular}


as c\caption{\textbf{Top-10 ranked recommendations/drugs predicted for SARS-Cov-2 by the DVA computational methods}.}
\label{t:top10-covid}
\end{table}


As can be observed, Table \ref{t:top10-covid} reports the drug recommendations for the novel Corona virus (nCov/SARS-CoV-2) using the proposed algorithm and other baselines.
The results from our proposed algorithm are sensible. Ribavarin is an FDA approved antiviral being considered as a treatment for COVID19 either by itself \cite{khalili2020novel} or as a part of combination \cite{hung2020triple}.
Remdesivir is a promising candidate for treating COVID19 having shown efficacy against SARS-CoV-1. The antiviral has been approved for emergency use by the US FDA \cite{eastman2020remdesivir,sarpatwari2020missed}; it has been approved by the drug controller general of India (DGCI) to treat mild to moderate COVID-19 patients 1 \footnote[1]{https://www.thehindu.com/sci-tech/health/indias-drug-regulator-grants-gilead-sciences-marketing-authorisation-
for-remdesivir/article31727034.ece}. The science behind the mechanism of Remdesivir’s action for inhibiting COVID19 has been thoroughly studied \cite{yin2020structural}.
\cite{yang2020effectiveness} showed that Umifenovir (brand name Arbidol) is effective as a prophyalactic against COVID19 infection. It has also found to be effective in the treatment of COVID affected subjects \cite{wang2020anti,wang2020anti,vankadari2020arbidol,xu2020arbidol}. DGCI has approved it for phase 3 trials in India \footnote[2]{https://futuremedicineindia.com/umifenovir-gets-nod-for-phase-iii-clinical-trial-against-covid-19-in-india/}. 
Favipiravir, a Russian drug for influenza, has also been undergoing trials in Japan; it has been sent to 40+ countries for COVID19
trials \footnote[3]{https://www.cnbc.com/2020/05/04/fujifilms-flu-drug-favipiravir-sent-to-43-nations-for-covid-19-trials.html}. \cite{shannon2020favipiravir} explains the scientific reason behind the prospective success of this antiviral. Baloxavir Marboxil has undergone trial in China for treating SARS-COV2 \footnote[4]{http://www.chictr.org.cn/showprojen.aspx?proj=49015}. Interferon alfa-2a in combination with Umifenovir have been found to be effective against COVID19 infected pneumonia \cite{xu2020arbidol}. In silico docking studies shows the potential of Baloxavir marboxil \cite{sahoo2020computational}; \cite{lipsitch2020testing} suggests that this antiviral can be effective if administered within 24 hours of the onset of infection. Several in silico docking studies have shown Dolutegravir may be effective in treating COVID19 infection \cite{beg2020anti,lee2020computational}. Trials for Chloroquine and alternately hydroxychloroquine was once suspended by WHO based on a controversial article, but they were
resumed within a few days of suspension \footnote[5]{https://www.newsbytesapp.com/timeline/world/61795/289490/who-set-to-resume-hydroxychloroquine-trials}. This antimalarial drug has been found to be effective against corona in multiple studies \cite{duan2020trial}; the scientific reason behind the efficacy of this drug against COVID19 has been explained in \cite{hu2020insights}. Geldanamycin and Laninamivir, predicted by our algorithm has not been considered for treating COVID19 patients.

The benchmark techniques GRMC, GRBMC and GRMF have three drugs common to ours – Ribavarin,
Remdesivir and Umifenovir. These three techniques also predict an anti-hepatitis C drug – Sofosbuvir; this drug was known to be effective against the coronavirus family \cite{elfiky2017quantitative}; it’s mechanism of action against COVID19 has also been studied \cite{elfiky2020ribavirin}. Evidence from clinical studies have corroborated the same \cite{jacome2020sofosbuvir}. GRMC and GRBMC also predicts chloroquine (as do we); but
GRMF is unable to predict it.

Note that the results shown in Table \ref{t:top10-covid} are ranked. The ranking is important, since it gives us a faith in the efficacy of the drugs. Our method predicts 6 drugs under trial among the top 7. The benchmarks only recommend 3 drugs in correct order. In fact, had we looked at the top 5 recommendations, only our algorithm would have recommended all of them correctly.

The benchmark algorithms predict three common drugs – Telaprevir \cite{shah2020silico} , Boceprevir \cite{fu2020both} and Simeprevir \cite{lo2020simeprevir,trezza2020integrated}. These drugs have the potential to be considered for COVID19 clinical trials. All the three benchmarks predict the antibody Palivizumab; there is no evidence in medical literature to support its usage for treating COVID19 patients or using it as a prophylactic. They also predict Taribavirin, which is similar to Ribavirin, and is used for treating Hepatitis-C; it has not been considered for clinical trials for SARS-COV2.

GRMF predicts a unique drug – Paritaprevir. An in silico docking study \cite{shah2020silico} showed that it has the potential for treating COVID19 much like Telaprevir, Boceprevir and Simeprevir.

The summary of the findings are in the following Table \ref{t:top10-covidsummary}.
\begin{table}[h!]
\centering
\begin{tabular}{l|l|l|l}
\hline \hline
Recommendations & \vtop{\hbox{\strut \#predictions used}\hbox{\strut in trial}} & \vtop{\hbox{\strut \#predictions}\hbox{\strut having potential}}& \vtop{\hbox{\strut \#predictions not}\hbox{\strut used/discontinued}} \\
\hline 
\hline		

GRDMF-2L &6&2&2\\
\hline
GRMC&5&3&2\\
\hline
GRBMC&5&3&2\\
\hline
GRMF&4&4&2\\
\hline 
\hline
\end{tabular}
\caption{\textbf{Top-10 recommendations/drugs predicted for SARS-Cov-2 by the DVA computational methods}.}
\label{t:top10-covidsummary}
\end{table}

Only our proposed algorithm predicts 6 drugs that are under trial (Remdesivir, Ribavarin, Umifenovir, Chloroquine, Favipiravir and Interferon alfa-2a). Two of the benchmarks predict 5 drugs that are under trial (Remdesivir, Ribavarin, Umifenovir, Sofosbuvir and Chloroquine); one of them (GRMF) misses out on Chloroquine. 

It is encouraging to observe that all the techniques (proposed and benchmarks) are predicting common drugs – Remdesivir, Ribavarin, Umifenovir.




\section{Conclusion}
AI researchers all across the world are contributing towards fighting COVID19. This is our humble contribution towards the same goal \cite{luengo2020artificial,hu2020challenges}. In this paper, we have proposed a novel optimization approach to solve the drug-virus association prediction problem. The contribution of this work is two-fold. Biologically, we have extended the phenomena of solving drug-virus association prediction for drug re-positioning using multiple similarities (graphs) and introduced new similarity measures for both drugs (based on mechanism of action) and viruses (based on symptoms). Technically, the proposed novel technique brings together the benefits of deep learning (using deep matrix factorization) and multiple graph learning (by regularization with multiple graphs). The algorithmic solution to the problem, relying on the novel HyPALM approach, does not require the setting of any extra hyper-parameter and has sounded convergence guarantees. We have made the software publicly available at \url{https://github.com/aanchalMongia/DeepVir}.

\section{Methods}
\subsection{Background}
\subsubsection{Modeling Biological Interactions}
A significant amount of research has been done in formulating drug re-positioning as a machine learning task. Various approaches such as network diffusion, supervised classification, neighborhood based prediction, clustering and matrix completion have been used in the past to address the said problem. Of these matrix completion techniques have turned out to be the most promising approach \cite{survey1}. Various biological applications which model the problem of drug-re-positioning using matrix completion include:
\begin{itemize}
    \item \textbf{Drug target interaction prediction:} The problem is to predict interaction score between a drug and target protein in a matrix assumed to have partial interaction data with drugs on rows and proteins on columns \cite{liu2016neighborhood,ezzat2017drug,eslami2018dmf,wang2018drug,mongia2020drug,zheng2020predicting}.
    \item \textbf{Drug Disease association prediction:} One predicts the probability that a certain drug will interact with a disease or not in a partially filled drug-disease association matrix. \cite{zhang2018predicting,luo2018computational,yang2019drug,mongia2020computationalb}.
    \item \textbf{Drug-Drug interaction prediction:} The drug-drug relationship is detected through a symmetrical drug-drug matrix/network  \cite{zhang2018manifold}. This helps predict drugs similar to the ones known to be effective against a pathogen/disease. Attempts have also been made to propose a machine learning approach for predicting drug-likeness of a molecule to distinguish potential drugs from small molecules that lack drug-like features \cite{beker2020minimal}.
\end{itemize}

In addition to drug re-positioning, other biological interaction problems have also be modeled  as matrix completion with the common goal of discovering or improving treatment of human diseases, for example we can cite:
\begin{itemize}
    \item \textbf{Gene disease association prediction:} Disease and gene features are used to learn genes for diseases as a gene-disease association prediction \cite{natarajan2014inductive,wu2008network,kohler2008walking}. This assists deciphering genetic basis of human diseases and better understanding of gene functions, interactions, and pathways.
    \item \textbf{Protein-protein interaction prediction:} Interactions are estimated in a protein-protein network \cite{wei2010low,kshirsagar2017multitask}. Kshirsagar \emph{et al.} \cite{kshirsagar2017multitask} deploy a matrix completion variant to model interactions between host (human here) proteins and pathogen (viruses causing infectious diseases here) proteins. This helps identifying the interaction between viral proteins and the human proteins, enabling deeper understanding of infectious diseases (which may involve biologically similar pathogens).
\end{itemize}

Motivated by such works, the problem of Drug virus association prediction was considered in~\cite{mongia2020computational}. Although solvable by any computational approach, the authors propose to tackle it using matrix completion motivated by the success of matrix completion in the above-mentioned applications. 

Among all the methodologies compared in \cite{mongia2020computational}, graph regularized matrix factorization based technique (GRMF) provided the best results for the validation setting where drugs are predicted for novel viruses. This impels us to come up with an improved algorithm built upon GRMF and leverage the benefit of deep learning and various kinds of metadata associated with drugs/viruses into the current method to aid drug re-positioning for viruses (including SARS-CoV-2).

\subsubsection{Matrix completion}
The goal of matrix completion is to recover all the entries of a matrix given a subset of known entries. Assume that $Y$ is a masked / undersampled version of the complete matrix $X$. Then, the problem can be expressed as:
\begin{equation}\label{eq:1}
Y = M \odot X,
\end{equation}

Here, $M$ denotes the mask which is element-wise multiplied to $X$ and has ones at positions where the values are known and zeros elsewhere. In general, the problem is under-determined with infinitely many solutions. However, when the sought matrix $X$ is of low-rank, several efficient resolution methods exist. The simplest of them is the matrix factorization approach \cite{cobanoglu2013predicting,ezzat2017drug}. Although known to work empirically, a theoretical understanding of factorization based completion is relatively recent \cite{sun2016guaranteed}. In this approach to matrix completion, $X$ is factored into a product of two low-rank matrices $U$ and $V$. Formally, this is expressed as:
\begin{equation}\label{eq:1}
Y = M\odot (UV).
\end{equation}
The factor matrices ($U$ and $V$) are obtained  by minimizing the following objective function:
\begin{equation}\label{eq:1}
\operatorname{minimize}_{U,V}
||Y - M\odot (UV)||_F^2,
\end{equation}
with $\| \cdot \|_F$ the Frobenius norm. The setting of the size of the latent factors $U$ and $V$ allow to impose rank constraints on their product.

Motivated by the success of deep learning in numerous fields \cite{shen2017deep,socher2012deep,schramowski2020making,mongia2019deepmc}, the shallow (two factor) models have been extended to deeper versions. The general factorization problem when the matrix is completely observed have been proposed in  \cite{li2017weakly,trigeorgis2017deep}. The solution to matrix completion via deep factorization has been proposed very recently, in \cite{mongia2019deepmc}. This can be formally expressed as follows, in the 2-layers case: 
\begin{equation}\label{eq:1}
\operatorname{minimize}_{U_1,U_2,V}
||Y - M\odot (U_1 U_2 V)||_F^2.
\end{equation}

Hereagain, the size of the latent factors $U_1$, $U_2$ and $V$ control the rank of the resulting product. One can extent to deeper layer formulations (3-layers and 4-layers) by factorizing $X$ into 4 and 5 matrices respectively. The first factor matrix $U_1$ (of size $m \times k_1$) contains the  representation of the row elements while the last factor matrix $V$ (of size $k_2 \times n$) is associated to its column representations in the latent space. The values for $k_1$ and $k_2$ here represent the number of latent factors, and are typically set using cross-validation.

Matrix factorization yields non-convex optimization problems which may be difficult to solve. A more direct approach to low rank matrix completion is to directly estimate matrix $X$ under low-rank constraints. Since the minimization of the rank leads to NP hard problems, one usually replaces it by its convex proxy, the nuclear norm~ \cite{candes2009exact,candes2010power}, defined as the sum of singular values. The problem formulation is then given as follows:
\begin{equation}\label{eq:nnm}
\operatorname{minimize}_X 
||Y - M\odot X||_F^2+||X||_*.
\end{equation}

In regular matrix completion, the entries in the unknown matrix are assumed to be real-valued. In communications and signal processing, this is actually never the case as those values are almost always quantized. This led to the  quantized matrix completion problem \cite{esmaeili2019novel,bhaskar2016probabilistic}. One extreme case is binary matrix completion \cite{davenport20141} where the matrix entries are represented by only one bit. Binary matrix completion is particularly appropriate for modeling biological interactions, so unsurprisingly studies in the past have used it for microbe disease association \cite{shi2018bmcmda}.

In all the aforesaid approaches, there was no scope for incorporating associated metadata, to constrain the completion problem and reduce the indeterminacy. In problems of biological imputations, there are often several sources of associated metadata. For example, in drug target interaction, one can compute the SIMCOMP score between the drugs based on their chemical structure as a drug-drug similarity measure \cite{hattori2010simcomp}. Similarly, similarities between target protein can be found a priori by computing the Smith-waterman score between the amino acid sequences of the target proteins \cite{amozidentification}. Studies have shown that the incorporation of associated metadata can improve the prediction results considerably \cite{ding2013similarity}. There can be several approaches to model the associated metadata, in the framework of matrix completion. One of the most promising approaches is to model the associations as graphs. This led to weighted graph regularised matrix factorization \cite{ezzat2017drug}. The formulation was expressed as:
\begin{equation}
\begin{array}{l}
 \label{eq:mgrdmf1}
 \operatorname{minimize}_{U,V} ||Y - M\odot (UV)||_F^2 + 2{\mu} \operatorname{tr}({U^\top}  {L_{r}} U) + 2{\mu}\operatorname{tr}(V  {L_c} {V^\top}). \hfill 
\end{array}
\end{equation}

Hereabove, $\mu \geq 0$ is the parameter penalizing the graph regularization terms, and $\operatorname{tr}(\cdot)$ denotes the trace of a matrix. Furthermore, the cost function in  \eqref{eq:mgrdmf1} incorporates Laplacian weight matrices $L_{r}$ and $L_c$ imposing prior correlation between row entities and column entities, respectively. Those matrices are obtained from known similarity factors $S_{r}$ and $S_c$, according to
\begin{equation}
    L_{r} = D_{r} - S_{r}, \quad L_{c} = D_{c} - S_{c}.
\end{equation}
$D_{r}$ and $D_c$ are diagonal degree matrices for the corresponding similarity matrices $S_{r}$ or $S_{c}$, that is, their $i$-th diagonal entry equals $D_{r}^{ii}=\sum_j S_{r}^{ij}$ and $D_{c}^{ii} = \sum_j S_{c}^{ij}$. 

We recently proposed a deep formulation for  \eqref{eq:mgrdmf1} in \cite{mongia2020deep} which amounts to solving: 
\begin{equation}
\begin{array}{l}
 \label{eq:mgrdmf1}
 \operatorname{minimize}_{U_1,U_2,V}  ||Y - M \odot (U_ 1 U_ 2 V)||_F^2 + {\mu _1}\operatorname{tr}(U_1^\top  {L_r} U_1) + {\mu _2}\operatorname{tr}(V  {L_c} {V^\top}).
\end{array}
\end{equation}
Graph regularised version of nuclear norm minimization was also proposed in \cite{kalofolias2014matrix} and was employed for drug target interactions \cite{wang2018drug}. The formulation is given by:
\begin{equation} \label{eq:grnnm}
\begin{array}{l}
\operatorname{minimize}_X  ||Y - M\odot X||_F^2 + \lambda ||X|{|_*} + {\mu _1}\operatorname{tr}({X^\top} {L_r} X) +
{\mu _2}\operatorname{tr}(X{L_c} {X^\top})
\end{array}
\end{equation}
All the aforementioned graph regularised techniques have been compared in our prior work \cite{mongia2020computational} in the context of drug-virus association prediction.

\subsection{Problem Formulation}
In this work, we assume the unknown matrix $X$ to be a complete drug-virus association matrix, with drugs along the rows ($m$ drugs) and viruses along the columns ($n$ viruses). It is a binary matrix with each entry $x_{ij}$ either equal to zero, representing no proven association between the corresponding drug and virus or to one, denoting that the $i^{th}$ drug is known to be effective against the $j^{th}$ virus. We set $Y$ as the observed matrix, corresponding to a sampled version from $X$. The information of the known association positions is encoded in the masking binary operator $M$. 

In our proposed technique, multi-graph regularization is incorporated in the deep matrix factorization formulation with the aim to incorporate the metadata associated with the drugs and viruses in the form of similarity information as shown below:
\begin{equation}
\begin{array}{l}
 \label{eq:mgrdmf1}
 \operatorname{minimize}_{U_1,U_2,V} ||Y - M\odot (U_1 U_ 2 V)||_F^2 + 2{\mu}\operatorname{tr}(U_1^\top  L_d U_1) + 2{\mu}\operatorname{tr}(V L_v {V^\top}), \quad \text{s.t.} \quad U_1 U_ 2 V \geq 0. \hfill 
\end{array}
\end{equation}
Hereabove, 
\begin{equation}
    L_d =\sum_{\ell=1}^{n_d} L_d^{\ell}, \quad L_v = \sum_{\ell=1}^{n_v} L_v^{\ell}.\label{eq:laplace}
\end{equation}
with $n_d$ and $n_v$ the number of available similarity matrices $(S_d^{\ell})_{\ell}$ and $(S_v^{\ell})_{\ell}$, for row entities (i.e. drugs) and column entities (i.e., viruses). The sizes of matrices $U_1, U_2$ and $V$ are $m\times k_1, k_1 \times k_2$ and $k_2 \times n$ respectively, where $k_1 \geq 1$ and $k_2 \geq 1$ are the numbers of latent factors assumed to be involved. Note that our formulation relies on the observation that computing the Laplacians associated to the sums $\sum_{\ell=1}^{n_d} S_d^{\ell}$ and $\sum_{\ell=1}^{n_v} S_v^{\ell}$ is actually equals to the sum of all the individual Laplacians involved \cite{mongia2020drug}, given by \eqref{eq:laplace}. 

In our previous work on drug-virus association prediction \cite{mongia2020computational}, we only consider one kind of similarity for drugs (based on the chemical structure of the drugs) and viruses (based on the genomic sequence of the viruses), so that $n_d$ and $n_v$ were equal to one. In this work, we propose to integrate a second type of similarities for drugs and for viruses by taking into account the mechanism of action of drugs and symptoms of viruses, so that $n_d=n_v=2$.  

The sought drug-virus association matrix $X$ to be recovered is assumed to be a positive entry matrix with a low-rank structure as similar drugs are known to affect biological systems (target pathways) in a similar fashion by having a similar mechanism of action \cite{jadamba2016systematic}, which motivates its formulation as the product of rectangular matrices, $X = U_1 U_2 V$ and the addition of the positivity constraint on it. 

\subsection{Data preprocessing}
We deploy the sparsification method to process the similarity matrices (as done in \cite{mongia2020drug,ezzat2017drug}) using $p$-nearest neighbor graph which is obtained by taking into account only the similarity values which correspond to the nearest neighbors for each drug/virus. $p$ here is set by performing cross-validation on the training set.

\subsection{Hybrid Proximal Alternating Linearized Minimization}
The minimization problem  \eqref{eq:mgrdmf1} is non-convex. The multi-linear structure of the involved operators motivates us for using the alternating minimization algorithm from \cite{RepettiJOGO}. This algorithm has been shown to be very powerful, in several applications such as seismic signal recovery \cite{SOOT},  phase retrieval \cite{RepettiPhase} and hyperspectral imaging \cite{RepettiUnmixing}. It can be seen as an alternating Majorization-Minimization (MM) technique~\cite{sun2017majorization}, including the powerful proximity operator \cite{Combettes} which allows to stabilize and speed up the convergence. In order to apply this method, one needs to introduce explicitly the variable $X$ in the problem formulation, as follows:
\begin{equation}
 \label{eq:mgrdmf2}
 \operatorname{minimize}_{X,U_1,U_2,V}  F(X,U_1,U_2,V)  \quad \text{s.t.} \quad X \geq 0,
\end{equation}
with
\begin{equation}
    F(X,U_1,U_2,V) = ||Y - M\odot X||_F^2 + \vartheta ||X - U_1 U_2 V||_F^2 + 2{\mu}\operatorname{tr}(U_1^\top  L_d U_1) + 2{\mu}\operatorname{tr}(V L_v {V^\top}).
\end{equation}
The advantage is to remove the complicated nonlinear constraint $U_1 U_2 V \geq 0$ while not deteriorating the solution (for $\vartheta>0$ sufficiently large). Function $F$ is then the sum of convex terms involving only a single parameter, and a quadratic, thus differentiable, non convex coupling term weighted by $\vartheta$. We are thus in the framework depicted in \cite{RepettiJOGO} (see also \cite{bolte2014proximal} for a particular case with two variables), so that we can apply the block coordinate variable metric forward-backward to solve Problem \eqref{eq:mgrdmf2}. This amounts to updating the four variables $X$, $U_1$, $U_2$ and $V$ sequentially, combining gradient (possibly with preconditioning) and proximal updates (see \cite{Combettes} for more details about the proximity operator). As emphasized in \cite{Abboud2014}, the two first quadratic terms can actually be processed either through proximal or gradient descent updates, without altering the convergence of the method. This nice flexibility allows us to propose the following alternating hybrid scheme, that we called HyPALM for the resolution of \eqref{eq:mgrdmf2}:
\begin{itemize}
    \item (update of $X$)
    Here, we perform a gradient step over the first quadratic term of \eqref{eq:mgrdmf2}, followed by a proximal step over the remaining terms:
    \begin{align}
        B & \leftarrow
        X - \alpha M\odot(M\odot X - Y),\\
        X & \leftarrow \operatorname{minimize}_Z \vartheta \|Z - U_1 U_2 V \|_F^2 + \varsigma \|Z - B \|_F^2  \quad \text{s.t.} \quad Z \geq 0,\\
        & \quad = \max \left \{ \frac{\varsigma B + \vartheta U_1 U_2 V}{\varsigma + \vartheta},0 \right \} \label{eq:max}.
    \end{align}
    Hereabove, $\varsigma>0$ is a parameter of the algorithm, and $ \alpha$ is a stepsize that must be chosen within the interval $]0,2[$ to secure the convergence of the method. The maximum operation in \eqref{eq:max} is taken entrywise, so it is equivalent to a simple cropping of the negative values.  
 \item (update of $U_1$) We solve the following minimization problem, equivalent to evaluate the proximity operator of $F$ restricted to the variable $U_1$, at the previous version of $U_1$ denoted $\overline{U_1}$:
 \begin{align}
     U_1 & \leftarrow \operatorname{argmin}_{U_1} F(X,U_1,U_2,V) + \varsigma \| U_1 - \overline{U_1}\|_F^2,\\
     & \quad = \operatorname{argmin}_{U_1} \vartheta ||X - U_1 U_2 V||_F^2 + 2{\mu}\operatorname{tr}(U_1^\top  L_d U_1) + \varsigma \| U_1 - \overline{U_1}\|_F^2.
 \end{align}
    By taking the gradient of the above expression with respect to $U_1$ equals to zero, we obtain:
    \begin{equation}
    \label{eq:sylvU1}
        (2 \mu L_{d} +\varsigma I){U_1} + \vartheta { U_1} (U_2 V)(U_2 V)^\top = X (U_2 V)^\top + \varsigma \overline{U_1}, 
    \end{equation}
    with $I$ the identity matrix. System \eqref{eq:sylvU1} takes the form of a Sylvester equation for which efficient solvers are available.
\item (update of $U_2$) Similarly to the latter update, we compute:
\begin{align}
    U_2 & \leftarrow \operatorname{argmin}_{U_2} F(X,U_1,U_2,V) + \varsigma \| U_2 - \overline{U_2}\|_F^2,\\
    & \quad = \operatorname{argmin}_{U_2} \vartheta ||X - U_1 U_2 V||_F^2 + \varsigma \| U_2 - \overline{U_2}\|_F^2,
\end{align}
with $\overline{U_2}$ the previous version of $U_2$. This leads to the resolution of the Sylvester equation:
\begin{equation}
        \vartheta U_2 V V^\top + \varsigma (U_1^\top U_1)^{-1} U_2 = \vartheta (U_1^\top U_1)^{-1} U_1^\top X V + \varsigma (U_1^\top U_1)^{-1} \overline{U_2}.
\end{equation}
\item (update of $V$) Hereagain, we solve the minimization problem:
\begin{align}
    V & \leftarrow \operatorname{argmin}_{V} F(X,U_1,U_2,V) + \varsigma \| V - \overline{V}\|_F^2,\\
    & \quad = \operatorname{argmin}_{V} \vartheta ||X - U_1 U_2 V||_F^2 +  2{\mu}\operatorname{tr}(V L_v {V^\top}) + \varsigma \| V - \overline{V}\|_F^2,
\end{align}
with $\overline{V}$ the previous version of $V$. This leads to solving the Sylvester equation:
\begin{equation}
 \vartheta (U_1 U_2)^\top U_1 U_2 V + V (2 \mu L_v + \varsigma I) = \vartheta (U_1 U_2)^\top X + \varsigma \overline{V}. 
\end{equation}
\end{itemize}

Iterating over the above four steps, with the simplified setting $ \varsigma = 1$, leads to Algorithm \ref{alg:dlfm_ppxa}, where we denote $\text{sylv}\{A,B,C\}$ the solution of the Sylvester equation $AX + XB = C$. The convergence of the iterates to a stationary point of Problem~\eqref{eq:mgrdmf2} is guaranteed as a consequence of the theoretical analysis in~\cite{RepettiJOGO,Abboud2014}.

\begin{algorithm}
\caption{HyPALM}\label{alg:dlfm_ppxa}
\small
\begin{algorithmic}[1]
\State \textbf{Set}: $ \mu , k_1, k_2, \alpha, L_d, L_v, \vartheta, K$
\State \textbf{Initialize} $X^{(0)}=Y$ and $(U_1^{(0)}, U_2^{(0)}, V^{(0)})$ (using SVD initialization)
\State \textbf{For} $k=1,2,\ldots,K$
\State  
\State $\;$ $X^{(k)} = \max \left\{\frac{B + \vartheta U_1^{(k - 1)} U_2^{(k - 1)} V^{(k - 1)}}{1 + \vartheta},0 \right\}$ with $B = X^{(k - 1)}  + \alpha ({M \odot} (Y - M \odot (X^{(k - 1)}))) $

\State  
\State  $\;$ $U_1^{(k)} = \text{sylv}\{2\mu L_d + I  ,   \vartheta U_2^{(k-1)} V^{(k-1)}(U_2 ^{(k-1)} V^{(k-1)})^\top,
\vartheta X^{(k)} 
(U_2^{(k-1)} V^{(k-1)})^\top 
+ U_1^{(k-1)}\}$ 
\State 
\State $\;$ $U_2^{(k)} = \text{sylv}\{((U_1^{(k)})^\top U_1^{(k)})^{-1} , \vartheta V^{(k-1)} (V^{(k-1)})^\top ,((U_1^{(k)})^\top U_1^{(k)})^{-1} U_2^{(k-1)} + \vartheta ((U_1^{(k)})^\top U_1^{(k)})^{-1} (U_1^{(k)})^\top X^{(k)} (V^{(k-1)})^\top   \}$
\State
\State $\;$ $V^{(k)} = \text{sylv}\{\vartheta (U_1^{(k)} U_2^{(k)})^\top U_1^{(k)} U_2^{(k)}, 2\mu L_v + I,  \vartheta (U_1^{(k)} U_2^{(k)})^\top X^{(k)} + V^{(k-1)}\}$
\State 
\State 
\State \textbf{End For}
\State \textbf{Return}: $X^{(K)}$
\end{algorithmic}
\end{algorithm}

It should be noted that the solution for deeper layers (3-layers, 4-layers, etc) of HyPALM method for the GRDMF problem can easily be obtained in a similar fashion to the 2-layers framework shown above.

\subsection{Similarity computation}
In this subsection, we describe in detail how each of the new similarities between drugs and viruses is obtained.

For drugs, we relied on the chemical structure-based SIMCOMP scores \cite{hattori2010simcomp} to represent drug-drug similarity as in \cite{mongia2020computational}. We also integrate an additional type of similarity by finding cosine similarities between one hot encoded representation of drug class based on mechanism of action of the drug. 
Similarly, we used the genomic structure-based $d2^*$ distance based on ONF (Oligonucleotide frequency) measure \cite{ahlgren2017alignment} as similarities between the viruses. To obtain the second similarity measure, we calculate the cosine similarity between one hot encoded representations of symptomatic profiles of viruses i.e. symptoms caused by the virus.
Both the drug class information and the file encoding which symptoms are caused by the virus (metadata used to generate new similarity measures) are available as supplementary.
Hence, the drug similarity matrix (of size $86 \times 86$) and virus similarity matrix (of size $23 \times 23$) are fixed and encode the metadata available.


\end{document}


\maketitle

\section{Selection of hyperparameters}
As we mention in section 2.2 of the main manuscript, we set the values of the hyperparameters corresponding to the proposed model by performing cross-validation on the training set for each of the three cross-validation settings explained in section 2.2.
These include:
\begin{itemize}
\item $\vartheta$: This is the weighting coefficient of the non convex coupling term in equation (13) in the main manuscript.
    \item $p$: This is the number of nearest drug//viruses to take into account within the similaririty (and hence Laplacian) matrices as mentioned in section 4.3 of the main manuscript. 
    \item $\alpha:$ This parameter corresponds to the step-size in the gradient update of HyPALM employed in equation (14).
    \item $\mu$: This is the regularization parameter used as coefficient of the graph Laplacian in the proposed formulation (see equation (12)).
    \item $k_1,k_2 \text{ and/or } k_3:$ Those are the numbers of latent factors assumed to be involved in deep matrix factorization. They correspond to the sizes of matrices $U_1, U_2$ and $V$ ($m\times k_1, k_1 \times k_2$ and $k_2 \times n$) respectively. $m$ and $n$ are the number of drugs and viruses in the data.
\end{itemize}
The optimal values for the hyperparameters for 2-layers and 3-layers versions of our method are shown in Tables \ref{t:para_2L} and \ref{t:para_3L} respectively.

\begin{table}[h!]
\centering
\begin{tabular}{l|llll|ll}
\hline \hline
&\vartheta       &   $p$ &\alpha    &\mu  & $k_1$ & 
$k_2$ \\
\hline
\hline
CV1&            1& 	   2& 	0.05& 100 &	17&15  \\
\hline 
\hline
CV2&            10&   2& 	0.01& 50  &  20&15  \\
\hline 
\hline
CV3&            2& 	  5& 	0.1 & 10 &  17&10   \\
\hline
\hline
\end{tabular}
\caption{Hyperparameter values for the proposed model 2 layer version}
\label{t:para_2L}
\end{table}

\begin{table}[h!]
\centering
\begin{tabular}{l|llll|lll}
\hline \hline
&\vartheta     &   $p$    &\alpha    &\mu  & $k_1$ & $k_2$ & $k_3$\\
\hline
\hline
CV1&           1& 	5& 	1&  5    &	23&10&7\\
\hline 
\hline
CV2&           1&   5& 	1&  0.01 &  20&15&10  \\
\hline 
\hline
CV3&           2& 	5& 1.5&  5   &  23& 10&7\\
\hline
\hline
\end{tabular}
\caption{Hyperparameter values for the proposed model 3 layer version}
\label{t:para_3L}
\end{table}

                            